\documentclass[oneside,letterpaper,10pt,onecolumn]{article}

\usepackage{arxivstyle}
\usepackage{authblk}

\usepackage{microtype}
\usepackage{graphicx}%
\usepackage{multirow}%
\usepackage{amsmath,amssymb,amsfonts}%
\usepackage{amsthm}%
\usepackage{mathrsfs}%
\usepackage[title]{appendix}%
\usepackage{color}%
\usepackage{textcomp}%
\usepackage{manyfoot}%
\usepackage{booktabs}%
\usepackage{algorithm}%
\usepackage{algorithmicx}%
\usepackage{algpseudocode}%
\usepackage{listings}%
\usepackage{cancel}
\usepackage{mathtools}
\usepackage[sort,comma,numbers]{natbib}
\usepackage{tcolorbox}

\newcommand{\diff}{\mathop{}\!\mathrm{d}}
\newcommand{\expp}{\mathrm{e}}
\newcommand{\cond}{{\;|\;}}

\definecolor{mylightblue}{rgb}{1.0, 0.53, 0.0}
\definecolor{mydarkblue}{rgb}{0, 0.08, 0.6} 
\usepackage[pagebackref,breaklinks,colorlinks=true]{hyperref}
\hypersetup{citecolor=mydarkblue,linkcolor=mylightblue}

\usepackage[capitalize,noabbrev]{cleveref}

\usepackage{soul}

\definecolor{purple}{rgb}{0.58, 0.44, 0.86}
\definecolor{orange}{rgb}{1.0, 0.55, 0.0}

\usepackage[frozencache=true]{minted}

\makeatletter
\renewcommand\AB@affilsepx{, \protect\Affilfont}
\makeatother

\begin{document}

\title{Taming Diffusion Models for Image Restoration: A Review}


\author[1]{\bf Ziwei Luo}
\author[2]{\bf Fredrik K. Gustafsson}
\author[3]{\bf Zheng Zhao}
\author[1]{\bf Jens Sj{\"o}lund}
\author[1]{\bf Thomas B. Sch{\"o}n}

\affil[1]{Uppsala University}
\affil[2]{Karolinska Institutet}
\affil[3]{Link\"{o}ping University}


\thispagestyle{empty}
\addtocounter{page}{-1}

\maketitle

\vspace{1cm}

\noindent \textbf{Please cite this version:}

\noindent \textit {Luo, Ziwei, Fredrik Gustafsson, Zheng Zhao, Jens Sjölund, and Thomas B. Schön. "Taming diffusion models for image restoration: a review." Philosophical Transactions A 383, no. 2299 (2025): 20240358.}

\begin{center}
\begin{minipage}{0.9\textwidth}
\begin{minted}[fontsize=\small,
  breaklines=true,
  breakanywhere=true,
  breaksymbolleft={},    
  breaksymbolright={},
  obeytabs=true,
  tabsize=2]{bibtex}
@article{luo2025taming,
  title={Taming diffusion models for image restoration: a review},
  author={Luo, Ziwei and Gustafsson, Fredrik and Zhao, Zheng and Sj{\"o}lund, Jens and Sch{\"o}n, Thomas B.},
  journal={Philosophical Transactions of the Royal Society A: Mathematical, Physical and Engineering Sciences},
  volume={383},
  number={2299},
  pages={20240358},
  year={2025},
  doi={10.1098/rsta.2024.0358},
}
\end{minted}
\end{minipage}
\end{center}

\vspace{1cm}

\noindent \textbf{A note on the structure of the article:}

\noindent The arXiv version of this article preserves all the core technical content and structure of the published article, but additionally includes extra sidebars (or “blocks”) that provide extended derivations, supplemental background discussion, and pointers to further reading. These blocks appear in the sections where they are first referenced, typically presented in grey-shaded boxes. In short: content is consistent between versions, with the arXiv version enriched by auxiliary explanatory boxes and references.

\newpage

\maketitle
\thispagestyle{plain}
\setcounter{page}{1}

\begin{abstract}
Diffusion models have achieved remarkable progress in generative modelling, particularly in enhancing image quality to conform to human preferences. Recently, these models have also been applied to low-level computer vision for photo-realistic image restoration (IR) in tasks such as image denoising, deblurring, dehazing, etc. In this review paper, we introduce key constructions in diffusion models and survey contemporary techniques that make use of diffusion models in solving general IR tasks. Furthermore, we point out the main challenges and limitations of existing diffusion-based IR frameworks and provide potential directions for future work.

\end{abstract}


\section{Introduction}\label{sec:intro}

Image restoration (IR) is a long-standing and challenging research topic in computer vision, which generally has two high-level aims: 1) recover high-quality (HQ) images from their degraded low-quality (LQ) counterparts, and 2) eliminate undesired objects from specific scenes. The former includes tasks like image denoising~\cite{buades2005review,cheng2021nbnet} and deblurring~\cite{shan2008high,kupyn2019deblurgan}, while the latter contains tasks like rain/haze/snow removal~\cite{valanarasu2022transweather,ozdenizci2023restoring} and shadow removal~\cite{le2019shadow,guo2023shadowdiffusion}. Figure~\ref{fig:ir-tasks} showcases examples of these applications. To solve different IR problems, traditional methods require task-specific knowledge to model the degradation and perform restoration in the spatial or frequency domain by combining classical signal processing algorithms~\cite{rabiner1975theory,orfanidis1995introduction} with specific image-degradation parameters~\cite{kundur1996blind,you1999blind,danielyan2011bm3d}. More recently, numerous efforts have been made to train deep learning models on collected datasets to improve performance on different IR tasks~\cite{zhao2016loss,zhang2017beyond,li2018benchmarking,pathak2016context}. Most of them directly train neural networks on sets of paired LQ-HQ images with a reconstruction objective (e.g., $\ell_1$ or $\ell_2$ distances) as typical in supervised learning. While effective, this approach tends to produce over-smooth results, particularly in textures~\cite{zhang2017beyond,ledig2017photo}. Although this issue can be alleviated by including adversarial or perceptual losses~\cite{goodfellow2020generative,johnson2016perceptual}, the training then typically becomes unstable and the results often contain undesired artifacts or are inconsistent with the input images~\cite{ledig2017photo,wang2018esrgan,zhang2019ranksrgan,pan2020physics}.

Recently, generative diffusion models (DMs)~\cite{ho2020denoising,song2020score} have drawn increasing attention due to their stable training process and remarkable performance in producing realistic images and videos~\cite{saharia2022photorealistic,ho2022imagen,rombach2022high,croitoru2023diffusion,yang2023diffusion}. Inspired by them, numerous works have incorporated the diffusion process into various IR problems to obtain high-perceptual/photo-realistic results~\cite{kawar2022denoising,saharia2022image,ozdenizci2023restoring,luo2023image,lian2024equipping}. However, these methods exhibit considerable diversity and complexity across various domains and IR tasks, obscuring the shared foundations that are key to understanding and improving diffusion-based IR approaches. In light of this, our paper reviews the key concepts in diffusion models and then surveys trending techniques for applying them to IR tasks. More specifically, the fundamentals of diffusion models are introduced in Sec.~\ref{sec:gmdm}, in which we further elucidate the score-based stochastic differential equations (Score-SDEs) and then show the connections between denoising diffusion probabilistic models (DDPMs) and Score-SDEs. In addition, the conditional diffusion models (CDMs) are elaborated such that we can learn to guide the image generation, which is key in adapting diffusion models for general IR tasks. Several diffusion-based IR frameworks are then methodologically summarised in Sec.~\ref{sec:dm4ir}. In particular, we show how to leverage CDMs for IR from different perspectives including DDPM, Score-SDE, and their connections. The connection even yields a training-free approach for non-blind IR, i.e. for tasks with known degradation parameters. Lastly, we conclude the paper with a discussion of the remaining challenges and potential future work in Sec.~\ref{sec:discussion}.

\section{Generative Modeling with Diffusion Models}\label{sec:gmdm}

Generative diffusion models (DMs) are a family of probabilistic models that tempers the data distribution into a reference distribution with an iterative process (e.g., Markov chains), and then learns to reverse this process for data sampling. In the following, Sec.~\ref{subsec:ddpms} describes a typical formulation of DMs: the denoising diffusion probabilistic models (DDPMs)~\citep{sohl2015deep,ho2020denoising}, followed by Sec.~\ref{subsec:score-sdes} that generalizes this to score-based stochastic differential equations (Score-SDEs) for a more detailed analysis of the diffusion/reverse process. Finally, in Sec.~\ref{subsec:guided_diffusion}, we further show how to guide DMs for conditional generation, which is a key enabling technique for diffusion-based IR.

\subsection{Denoising Diffusion Probabilistic Models (DDPMs)}\label{subsec:ddpms}

\begin{figure}[t]
    \centering
    \includegraphics[width=.95\linewidth]{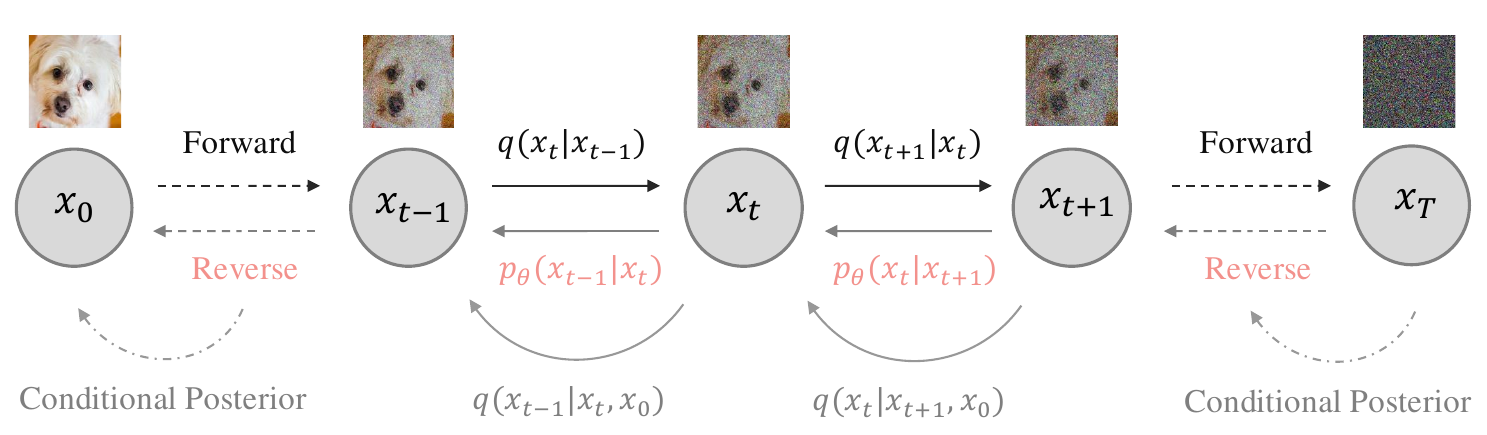}\vspace{-1.0mm}
    \caption{Denoising diffusion probabilistic models (DDPMs). The forward path transfers data to Gaussian noise, and the reverse path learns to generate data from noise along the actual time reversal of the forward process. Here, the reverse transition $p_{\theta}(x_{t-1} \mid x_t)$ represents the model we aim to learn, and the conditional posterior $q(x_{t-1} \mid x_t, x_0)$ is a tractable Gaussian which serves as the target distribution the model wants to match as the $L_{t-1}$ term in Eq.~\eqref{eq:diffusion_training_loss}.}\vspace{-1.0mm}
    \label{fig:ddpm}
\end{figure}

Given a variable $x_0$ sampled from a data distribution $q_0(x)$, DDPMs~\citep{sohl2015deep,ho2020denoising} are latent variable models consisting of two Markov chains: a forward/diffusion process $q(x_{1:T}\mid x_0)$ and a reverse process $p_\theta(x_{0:T})$. The forward process transfers $x_0$ to a Gaussian distribution by sequentially injecting noise. Then the reverse process learns to generate new data samples starting from the Gaussian noise. An overview of the DDPM is shown in Figure~\ref{fig:ddpm}. Below we elaborate on these two processes, and give details for how DDPMs are trained.

\subsubsection{Forward diffusion process} 
The forward process perturbs data samples $x_0$ to noise $x_T$. It can be characterized by a joint distribution encompassing all intermediate states, represented in the form:
\begin{equation}
    q(x_{1:T} \mid x_0) = \prod_{t=1}^T q(x_t \mid x_{t-1}), \quad x_0 \sim q_0(x),
    \label{eq:diffusion_forward}
\end{equation}
where the transition kernel $q(x_t \mid x_{t-1})$ is a handcrafted Gaussian given by
\begin{equation}
    q(x_t \mid x_{t-1}) = \mathcal{N}(x_t; \sqrt{1-\beta_t}x_{t-1}, \beta_t I),
    \label{eq:diffusion_forward_kernel}
\end{equation}
where $\beta_{1:T} \in (0, 1)$ is the variance schedule, a set of pre-defined hyper-parameters to ensure that the forward process (approximately) converges to a Gaussian distribution. Let $\alpha_t \coloneqq 1 - \beta_t$ and $\bar{\alpha}_t \coloneqq \prod_{s=1}^t \alpha_s$, Eq.~\eqref{eq:diffusion_forward_kernel} then allows us to marginalize the joint distribution of Eq.~\eqref{eq:diffusion_forward} to the following\footnote{Derivations can be found in \textbf{blocks} throughout the paper. If not interested in extra details, these blocks can safely be skipped.}:
\begin{equation}
    q(x_t \mid x_0) = \mathcal{N}(x_t; \sqrt{\bar{\alpha}_t} x_0, (1-\bar{\alpha}_t)I).
    \label{eq:diffusion_marginalize_kernel}
\end{equation}
{\small
\begin{tcolorbox}[title={\textbf{Block 1:} Derivation for the marginal distribution of Eq.~\eqref{eq:diffusion_marginalize_kernel}}]\label{block1}
\begin{proof}
By reparameterizing the forward transition kernel (Eq.~\eqref{eq:diffusion_forward_kernel}) with $\alpha_t$, we have 
\begin{align}
    x_t 
    &= \sqrt{\alpha_t}\,x_{t-1} + \sqrt{1-\alpha_t}\,\epsilon_{t-1}, \quad \quad \epsilon_{1:T} \sim \mathcal{N}(0, I) \nonumber \\
    &= \sqrt{\alpha_t \alpha_{t-1}}\,x_{t-2} + \underbrace{\sqrt{1-\alpha_t}\,\epsilon_{t-1} + \sqrt{\alpha_t(1-\alpha_{t-1})}\,\epsilon_{t-2}}_{= \sqrt{1-\alpha_t \alpha_{t-1}}\,\epsilon \quad \text{(sum of two Gaussian)}} \nonumber \\
    &= \dots \nonumber \\ 
    &= \sqrt{\alpha_t \alpha_{t-1} \cdots \alpha_1}\,x_0 + \sqrt{1-\alpha_t \alpha_{t-1} \cdots \alpha_1} \, \epsilon \nonumber \\
    &= \sqrt{\bar{\alpha}_t}\,x_0 + \sqrt{1-\bar{\alpha}_t} \, \epsilon. \nonumber
\end{align}
Meaning that $q(x_t \mid x_0) = \mathcal{N}(x_t; \sqrt{\bar{\alpha}_t} x_0, (1-\bar{\alpha}_t)I)$, which completes the proof.
\end{proof}
\end{tcolorbox}
}%
We usually set $\beta_1 < \beta_2 < \dots < \beta_T$ such that $\alpha_1 > \alpha_2 > \dots > \alpha_T \approx 0$ and the terminal distribution $q(x_T) \approx \mathcal{N}(x_T; 0, I)$ thus is a standard Gaussian, which allows us to generate new data points by reversing the diffusion process starting from sampled Gaussian noise. Moreover, it is important to note that posteriors along the forward process are tractable when conditioned on $x_0$, i.e. $q(x_{t-1} \mid x_t, x_0)$ is a tractable Gaussian~\cite{sohl2015deep}. This tractability enables the derivation of the DDPM training objective, which we will describe in Sec.~\ref{subsubsec:ddpm_objective}.

\subsubsection{Reverse process} 
In contrast, the reverse process learns to match the actual time reversal of the forward process, which is also a joint distribution modelled by $p_\theta(x_{0:T})$ as follows:
\begin{equation}
    p_\theta(x_{0:T}) = p(x_T) \prod_{t=1}^T p_{\theta}(x_{t-1} \mid x_t), \quad x_T \sim \mathcal{N}(0, I).
    \label{eq:diffusion_reverse}
\end{equation}
In DDPMs, the transition kernel $p_{\theta}(x_{t-1} \mid x_t)$ is defined as a learnable Gaussian:
\begin{equation}
    p_{\theta}(x_{t-1} \mid x_t) = \mathcal{N}(x_{t-1}; \mu_{\theta}(x_t, t), \, \Sigma_\theta (x_t, t) I),
    \label{eq:diffusion_reverse_kernel}
\end{equation}
where $\mu_\theta$ and $\Sigma_\theta$ are the parameterised mean and variance, respectively. Then learning the model of Eq.~\eqref{eq:diffusion_reverse_kernel} is key to DDPMs since it substantially affects the quality of data sampling. That is, we have to adjust the parameters $\theta$ until the final sampled variable $x_0$ is close to that sampled from the real data distribution.

\subsubsection{Training objective}\label{subsubsec:ddpm_objective}

To learn the reverse process, we usually minimize the variational bound on the negative log-likelihood which introduces the forward joint distribution of Eq.~\eqref{eq:diffusion_forward} in the objective $L$ as (we simplify $\mathbb{E}_{q(x_{0:T})}$ as $\mathbb{E}_q$):
{
\begin{equation}
    \mathbb{E}_{q_0(x)} [-\log p_\theta(x_0) ] \leq \underbrace{\mathbb{E}_{q(x_{0:T})}\Bigl[ -\log \frac{p_\theta(x_{0:T})}{q(x_{1:T} \mid x_0)} \Bigr]}_{\text{negative evidence lower bound (ELBO)}} = \mathbb{E}_{q} \Bigl[ -\log p(x_T) - \sum_{t=1}^T \log \frac{p_\theta (x_{t-1} \mid x_t)}{q(x_t \mid x_{t-1})} \Bigr].
    \label{eq:diffusion_elbo}
\end{equation}
}%
Here, $p(x_T)$ is a standard Gaussian, $p_\theta (x_{t-1} \mid x_t)$ is the reverse transition kernel Eq.~\eqref{eq:diffusion_reverse_kernel} that we want to learn, and $q(x_t \mid x_{t-1})$ is the forward transition kernel Eq.~\eqref{eq:diffusion_forward_kernel}. This objective can be further rewritten to:
{\small
\begin{align}
    L \coloneqq \mathbb{E}_{q} \Bigl[ \underbrace{D_{KL}(q(x_T \mid x_0) \mid\mid p(x_T))}_{L_T} + \sum_{t=2}^{T} \underbrace{D_{KL}(q(x_{t-1} \mid x_t, x_0) \mid\mid p_\theta(x_{t-1} \mid x_t))}_{L_{t-1}} \underbrace{-\log p_\theta (x_0 \mid x_1)}_{L_0} \Bigr],
    \label{eq:diffusion_training_loss}
\end{align}
}%
where $L_T$ is called the prior matching term and contains no learnable parameters, $L_{t-1}$ is the posterior matching term, and $L_0$ the data reconstruction term that maximizes the likelihood of~$x_0$. \citet{sohl2015deep} have proved that the conditional posterior distribution in $L_{t-1}$ is a tractable Gaussian: $q(x_{t-1} \mid x_t, x_0) = \mathcal{N}(x_{t-1}; \tilde{\mu}_t(x_t, x_0), \tilde{\beta}_t I)$, where the mean and variance are given by
\begin{align}
    \tilde{\mu}_t(x_t, x_0) \coloneqq \frac{\sqrt{\alpha_t}(1 - \bar{\alpha}_{t-1})}{1 - \bar{\alpha}_t} x_t + \frac{\sqrt{\bar{\alpha}_{t-1}} \beta_t}{1 - \bar{\alpha}_t} x_0, \quad \text{and} \quad
    \tilde{\beta}_t \coloneqq \frac{1 - \bar{\alpha}_{t-1}}{1 - \bar{\alpha}_t} \beta_t,
    \label{eq:diffusion_conditional_posterior_solution}
\end{align}
All terms in $\tilde{\beta}_t$ are known and thus the posterior variance in~\eqref{eq:diffusion_reverse_kernel} can be nonparametric, i.e., $\Sigma_\theta (x_t, t) = \tilde{\beta}_t$.
Then, applying the reparameterization trick to $q(x_t \mid x_0)$ of Eq.~\eqref{eq:diffusion_marginalize_kernel} gives an estimate of the initial state: $x_0 = \frac{1}{\sqrt{\bar{\alpha}_t}}(x_t - \sqrt{1 - \bar{\alpha}_t}\epsilon_t)$, which can be substituted into Eq.~\eqref{eq:diffusion_conditional_posterior_solution} to obtain: $\tilde{\mu}_t(x_t, x_0) = \frac{1}{\sqrt{{\alpha}_t}}(x_t - \frac{1 - \alpha_t}{\sqrt{1 - \bar{\alpha}_t}} \epsilon_t)$.
The only unknown part here is the noise $\epsilon_t$ which can be learned using a neural network $\epsilon_\theta(x_t, t)$, and the parameterised distribution mean can be rewritten as:
\begin{equation}
    \mu_\theta(x_t, t) = \frac{1}{\sqrt{{\alpha}_t}}(x_t - \frac{1 - \alpha_t}{\sqrt{1 - \bar{\alpha}_t}} \epsilon_\theta(x_t, t)).
    \label{eq:diffusion_learnable_posterior_mean}
\end{equation}
The transition kernel $p_{\theta}(x_{t-1} \mid x_t)$ of Eq.~\eqref{eq:diffusion_reverse_kernel} is finally updated according to the following:
\begin{equation}
    p_{\theta}(x_{t-1} \mid x_t) = \mathcal{N}(x_{t-1}; \mu_\theta(x_t, t), \, \tilde{\beta}_t I).
    \label{eq:learnable_diffusion_reverse_kernel}
\end{equation}
Note that $p_{\theta}(x_{t-1} \mid x_t)$ now matches the form of $q(x_{t-1} \mid x_t, x_0) = \mathcal{N}(x_{t-1}; \tilde{\mu}_t(x_t, x_0), \tilde{\beta}_t I)$, in order to minimise the KL term of $L_{t-1}$ in Eq.~\eqref{eq:diffusion_training_loss}. Also note that DDPMs only need to learn the noise network $\epsilon_\theta(x_t, t)$, for which it is common to use a U-Net architecture with several self-attention layers~\cite{ho2020denoising}. The noise network $\epsilon_\theta(x_t, t)$ takes an image $x_t$ and a time $t$ as input, and outputs a noise image of the same shape as $x_t$. 
More specifically, the scalar time $t$ is encoded into vectors similar to the positional embedding~\cite{vaswani2017attention} and is combined with $x_t$ in the feature space for time-varying noise prediction. 
{\small
\begin{tcolorbox}[title={\textbf{Block 2:} Complete derivation for the training objective of Eq.~\eqref{eq:diffusion_training_loss}}]\label{block2}
Firstly, let's derive the diffusion objective for a single image $x_0$:
\begin{equation}
    \begin{aligned}
    \tilde{L} &\coloneqq -\log p_\theta(x_0) \\
    &= -\log \int p_\theta(x_{0:T}) \diff x_{1:T} \\
    &= -\log \int \frac{p_\theta(x_{0:T}) q(x_{1:T} \mid x_0)}{q(x_{1:T} \mid x_0)} \diff x_{1:T} \\
    &= -\log \mathbb{E}_{q(x_{1:T} \mid x_0)}\Bigl[ \frac{p_\theta(x_{0:T})}{q(x_{1:T} \mid x_0)} \Bigr] \\
    &\leq \underbrace{\mathbb{E}_{q(x_{1:T} \mid x_0)}\Bigl[ -\log \frac{p_\theta(x_{0:T})}{q(x_{1:T} \mid x_0)} \Bigr]}_{\textit{negative evidence lower bound (ELBO)}}  \quad \quad  \text{(Jensen's Inequality)} \\
    &= \mathbb{E}_{q(x_{1:T} \mid x_0)} \Bigl[ -\log \frac{p(x_T) \prod_{t=1}^T p_\theta (x_{t-1} \mid x_t)}{\prod_{t=1}^T q(x_t \mid x_{t-1})} \Bigr] \quad \quad \text{(Eq.~\eqref{eq:diffusion_reverse} and Eq.~\eqref{eq:diffusion_forward})} \\
    &= \mathbb{E}_{q(x_{1:T} \mid x_0)} \Bigl[ -\log p(x_T) - \sum_{t=2}^T \log \underbrace{\frac{p_\theta (x_{t-1} \mid x_t)}{q(x_{t-1} \mid x_t, x_0)} \, \frac{q(x_{t-1} \mid x_0)}{q(x_t \mid x_0)}}_{\text{Bayes' rule on} \; q(x_t \mid x_{t-1})} - \log \frac{p_\theta (x_0 \mid x_1)}{q(x_1 \mid x_0)} \Bigr] \\
    &= \mathbb{E}_{q(x_{1:T} \mid x_0)} \Bigl[ -\log \frac{p(x_T)}{q(x_T \mid x_0)} - \sum_{t=2}^T \log \frac{p_\theta (x_{t-1} \mid x_t)}{q(x_{t-1} \mid x_t, x_0)} - \log p_\theta (x_0 \mid x_1) \Bigr] \\
    &= D_{KL}(q(x_T \mid x_0) \mid\mid p(x_T))\\
    &\quad \quad + \sum_{t=2}^{T} \mathbb{E}_{q(x_t \mid x_0)} \Bigl[D_{KL}(q(x_{t-1} \mid x_t, x_0) \mid\mid p_\theta(x_{t-1} \mid x_t)) \Bigr] - \mathbb{E}_{q(x_1 \mid x_0)} \Bigl[\log p_\theta (x_0 \mid x_1) \bigr]. \nonumber
\end{aligned}
\end{equation}
Then, adding the expectation of sampled image $x_0$ to $\tilde{L}$ yields the final objective:
\begin{equation}
    L \coloneqq \mathbb{E}_{q} \Bigl[ \underbrace{D_{KL}(q(x_T \mid x_0) \mid\mid p(x_T))}_{L_T} + \sum_{t=2}^{T} \underbrace{D_{KL}(q(x_{t-1} \mid x_t, x_0) \mid\mid p_\theta(x_{t-1} \mid x_t))}_{L_{t-1}} \underbrace{-\log p_\theta (x_0 \mid x_1)}_{L_0} \Bigr]. \nonumber
\end{equation}
\end{tcolorbox}
}%
\vspace{-0.2in}
{\small
\begin{tcolorbox}[title={\textbf{Block 3:} Complete derivation for the conditional posterior distribution of Eq.~\eqref{eq:diffusion_conditional_posterior_solution}}]\label{block3}
With Bayes's rule, the conditional posterior distribution can be rewritten as:
\begin{equation}
\begin{aligned}
    q(x_{t-1} \mid x_t, x_0) 
    &= \frac{q(x_t \mid x_{t-1}) q(x_{t-1} \mid x_0))}{q(x_t \mid x_0)} \quad \quad \text{(Bayes' rule)} \\
    &= \frac{\mathcal{N}(x_t; \sqrt{\alpha_t}x_{t-1}, \beta_t I) \cdot \mathcal{N}(x_{t-1}; \sqrt{\bar{\alpha}_{t - 1}} x_0, (1 - \bar{\alpha}_{t-1}) I)}{\mathcal{N}(x_t; \sqrt{\bar{\alpha}_t} x_0, (1 - \bar{\alpha}_t) I)} \\
    &= (2\pi \beta_t)^{-\frac{d}{2}} \cdot (2\pi (1-\bar{\alpha}_{t-1}))^{-\frac{d}{2}} \cdot (2\pi (1-\bar{\alpha}_t))^{\frac{d}{2}} \\ 
    &\quad \cdot \exp \bigl(-\frac{\|x_t - \sqrt{\alpha_t}x_{t-1}\|^2}{2 \beta_t} - \frac{\|x_{t-1} - \sqrt{\bar{\alpha}_{t-1}}x_0 \|^2}{2 (1 - \bar{\alpha}_{t-1})} + \frac{\|x_t - \sqrt{\bar{\alpha}_t}x_0 \|^2}{2 (1 - \bar{\alpha}_t)} \bigr) \\
    &= (2\pi\frac{1 - \bar{\alpha}_{t-1}}{1 - \bar{\alpha}_t}\beta_t)^{-\frac{d}{2}} \cdot \exp \Bigl({-\frac{\| x_{t-1} - (\frac{\sqrt{\bar{\alpha}_{t-1}} \beta_t}{1 - \bar{\alpha}_t} x_0 + \frac{\sqrt{\alpha_t}(1 - \bar{\alpha}_{t-1})}{1 - \bar{\alpha}_t} x_t) \|^2}{\frac{1 - \bar{\alpha}_{t-1}}{1 - \bar{\alpha}_t} \beta_t}} \Bigr), \nonumber
\end{aligned}
\end{equation}
which is a Gaussian distribution:
\begin{equation}
    q(x_{t-1} \mid x_t, x_0) = \mathcal{N}(x_{t-1}; \frac{\sqrt{\alpha_t}(1 - \bar{\alpha}_{t-1})}{1 - \bar{\alpha}_t} x_t + \frac{\sqrt{\bar{\alpha}_{t-1}} \beta_t}{1 - \bar{\alpha}_t} x_0, \frac{1 - \bar{\alpha}_{t-1}}{1 - \bar{\alpha}_t}\beta_t I). \nonumber
\end{equation}
\end{tcolorbox}
}%

\paragraph{Simplified objective}
We now have known expressions for all components of the objective $L$ in Eq.~\eqref{eq:diffusion_training_loss}. Its current form is however not ideal to use for model training since it requires $L_t$ to be computed at every timestep of the entire diffusion process, which is time-consuming and impractical. Fortunately, the prior matching term $L_T$ can be ignored since it contains no parameters. By substituting Eq.~\eqref{eq:diffusion_conditional_posterior_solution} and \eqref{eq:diffusion_learnable_posterior_mean} into Eq.~\eqref{eq:diffusion_training_loss}, we also find that the final expanded version of the posterior matching term $L_{t-1}$ ($t \in \{2, \dots, T\}$) and data reconstruction term $L_0$ have similar forms, namely
\begin{equation}
    L_{t-1} \coloneqq \frac{\beta_t}{2 \alpha_t (1 - \bar{\alpha}_{t-1})} \mathbb{E}_{x_0, \epsilon} \Bigl[ \| \epsilon_t - \epsilon_\theta(x_t, t) \|^2 \Bigr] \quad \mathrm{and} \quad L_0 \coloneqq \frac{1}{2\alpha_1} \mathbb{E}_{x_0, \epsilon} \Bigl[\| \epsilon_1 - \epsilon_\theta(x_1, 1) \|^2 \Bigr].
    \label{eq:expanded_diffuion_training_loss}
\end{equation}
By ignoring the weights outside the expectations in Eq.~\eqref{eq:expanded_diffuion_training_loss}, a simplified training objective can therefore be obtained according to the following~\cite{ho2020denoising}:
\begin{align}
    L_\text{simple} 
    \coloneqq \mathbb{E}_{x_0, t, \epsilon} \Bigl[\| \epsilon_t - \epsilon_\theta(x_t, t) \|^2 \Bigr] 
    = \mathbb{E}_{x_0, t, \epsilon} \Bigl[\| \epsilon_t - \epsilon_\theta(\sqrt{\bar{\alpha}_t}x_0 + \sqrt{1-\bar{\alpha}_t}\cdot \epsilon, t) \|^2 \Bigr],
    \label{eq:simplified_diffusion_loss}
\end{align}
which essentially learns to match the predicted and real added noise for each training sample and thus is also called the \emph{noise matching loss}. Compared to the original objective $L$ in Eq.~\eqref{eq:diffusion_training_loss}, $L_\text{simple}$ is a re-weighted version that puts more focus on larger timesteps $t$, which empirically has been shown to improve the training~\cite{ho2020denoising}. Once trained, the noise prediction network $\epsilon_\theta(x_t, t)$ can be used to generate new data $x_0$ by running Eq.~\eqref{eq:learnable_diffusion_reverse_kernel} starting from $x_T \sim \mathcal{N}(0, I)$, i.e. by iterating
\begin{equation}
    x_{t-1} = \mu_\theta(x_t, t) + \sqrt{\tilde{\beta}_t}\epsilon     \quad \mathrm{where} \quad \mu_\theta(x_t, t) = \frac{1}{\sqrt{{\alpha}_t}}(x_t - \frac{1 - \alpha_t}{\sqrt{1 - \bar{\alpha}_t}} \epsilon_\theta(x_t, t)),
    \label{eq:diffusion_sample}
\end{equation}
as a parameterised data sampling process, similar to that in Langevin dynamics~\cite{song2019generative}.
{\small
\begin{tcolorbox}[title={\textbf{Block 4:} Complete derivation for the posterior matching term $L_{t-1}$ in Eq.~\eqref{eq:expanded_diffuion_training_loss}}]\label{block4}
First, let us recall the KL divergence of two Gaussian distributions:
\begin{equation}
    D_{KL}(p \mid\mid q) = \frac{1}{2} \Bigl[ (\mu_p - \mu_q)^\mathsf{T}\Sigma_q^{-1}(\mu_p - \mu_q) + \log \frac{|\Sigma_q|}{|\Sigma_p|} + tr\{ \Sigma_q^{-1}\Sigma_p \} - d \Bigr], \nonumber
\end{equation}
where $d$ is the data dimension. Then we can use this to compute the loss term $L_{t-1}$:\\
\begin{equation}
\begin{aligned}
    L_{t-1} 
    &\coloneqq \mathbb{E}_{q} \Bigl[ D_{KL}(\mathcal{N}(x_{t-1}; \tilde{\mu}_t(x_t, x_0), \tilde{\beta}_t I) \mid\mid \mathcal{N}(x_{t-1}; \mu_{\theta}(x_t, t), \tilde{\beta}_t I)) \Bigr] \\
    &= \frac{1}{2} \mathbb{E}_{x_0, \epsilon} \Bigl[ \frac{1}{\tilde{\beta}_t} \| \frac{1}{\sqrt{{\alpha}_t}}(x_t - \frac{1 - \alpha_t}{\sqrt{1 - \bar{\alpha}_t}} \epsilon_t) - \frac{1}{\sqrt{{\alpha}_t}}(x_t - \frac{1 - \alpha_t}{\sqrt{1 - \bar{\alpha}_t}} \epsilon_\theta(x_t, t)) \|^2 + \bcancel{tr\{I\} - d} \Bigr] \\
    &= \frac{1}{2 \cdot \tilde{\beta}_t} \mathbb{E}_{x_0, \epsilon} \Bigl[ \frac{(1 - \alpha_t)^2}{\alpha_t(1 - \bar{\alpha}_t)} \| \epsilon_t - \epsilon_\theta(x_t, t) \|^2 \Bigr] \\
    &= \frac{1}{2 \cdot \beta_t} \cdot \frac{1 - \bar{\alpha}_t}{1 - \bar{\alpha}_{t-1}} \cdot \frac{(1 - \alpha_t)^2}{\alpha_t(1 - \bar{\alpha}_t)} \mathbb{E}_{x_0, \epsilon} \Bigl[ \| \epsilon_t - \epsilon_\theta(x_t, t) \|^2 \Bigr] \\
    &= \frac{\beta_t}{2 \alpha_t (1 - \bar{\alpha}_{t-1})} \mathbb{E}_{x_0, \epsilon} \Bigl[ \| \epsilon_t - \epsilon_\theta(x_t, t) \|^2 \Bigr]. \nonumber
\end{aligned}
\end{equation}
\end{tcolorbox}
}%
{\small
\begin{tcolorbox}[title={\textbf{Block 5:} Complete derivation for the reconstruction term $L_0$ in Eq.~\eqref{eq:expanded_diffuion_training_loss}}]\label{block5}
\begin{equation}
\begin{aligned}
    L_0 
    &\coloneqq \mathbb{E}_q \bigl[-\log p_{\theta}(x_0 | x_1) \bigr] \\
    &= \mathbb{E}_{x_0, \epsilon} \Bigl[ -\log \mathcal{N}(x_0; \frac{1}{\sqrt{\alpha_1}}(x_1 - \frac{\beta_1}{\sqrt{1 - \alpha_1}} \epsilon_\theta(x_1, 1)), \beta_1 I) \Bigr] \\
    &= \frac{d}{2} \log 2\pi \beta_1 + \frac{1}{2\beta_1} \mathbb{E}_{x_0, \epsilon} \Bigl[ \frac{1 - \alpha_1}{\alpha_1} \| \epsilon_1 - \epsilon_\theta(x_1, 1) \|^2 \Bigr] \\
    &= \bcancel{\frac{d}{2} \log 2\pi \beta_1} + \frac{1}{2\alpha_1} \mathbb{E}_{x_0, \epsilon} \Bigl[\| \epsilon_1 - \epsilon_\theta(x_1, 1) \|^2 \Bigr]. \nonumber
\end{aligned}
\end{equation}
\end{tcolorbox}
}%

\subsection{Data Perturbation and Sampling with SDEs}\label{subsec:score-sdes}

\begin{figure}[t]
    \centering
    \includegraphics[width=.99\linewidth]{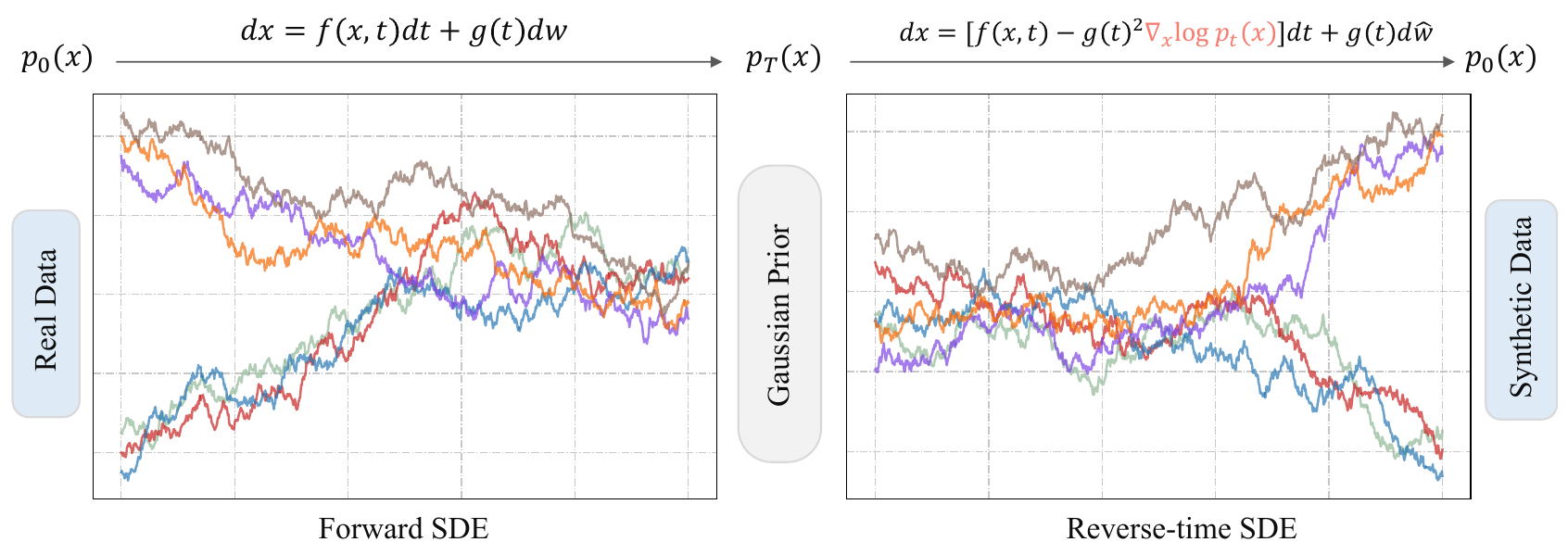}
    \caption{Data perturbation and sampling with SDEs. Different from DDPMs, Score-SDE continuously perturbs the data to Gaussian noise using a forward SDE, $\diff {x} = f(x, t) \diff t + g(t)\diff w$, and then generates new samples by estimating the score $\nabla_{{x}} \log p_t({x})$ and simulating the corresponding reverse-time SDE.}
    \label{fig:score-sde}
\end{figure}

We can further generalize the DDPM to stochastic differential equations, namely Score-SDE~\cite{song2020score}, where both the forward and reverse processes are in continuous-time state space. This generalization offers a deeper insight into the mathematics behind DMs that underlies the success of diffusion-based generative modelling. Figure~\ref{fig:score-sde} shows an overview of the Score-SDE approach.

\subsubsection{Data perturbation with forward SDEs}
Here, assume the real data distribution is $p_0(x)$, we construct variables $\{ x(t)\}_{t=0}^T$ for data perturbation in continuous time, which can be modeled as a forward SDE defined by
\begin{equation}
	\diff {x} = f(x, t) \diff t + g(t)\diff w, \quad x(0) \sim p_0(x), 
	\label{eq:forward-sde}
\end{equation}
where $f(x, t)$ and $g(t)$ are called the \textit{drift} and \textit{diffusion} functions, respectively, and $w$ is a standard Wiener process (a.k.a., Brownian motion). We use $p_t(x)$ to denote the marginal probability density of $x(t)$, and use $p\left(x(t) \cond x(s)\right)$ to denote the transition kernel from $x(s)$ to $x(t)$. Moreover, we always design the SDE to drift to a fixed prior distribution (e.g., standard Gaussian), ensuring that $x(T)$ becomes independent of $p_0(x)$ and can be sampled individually.

\subsubsection{Sampling with reverse-time SDEs}

We can sample noise and reverse the forward SDE to generate new data close to that sampled from the real data distribution. Note that reversing Eq.~\eqref{eq:forward-sde} yields another diffusion process, i.e. a reverse-time SDE~\cite{anderson1982reverse}:
\begin{equation}
    \diff {x} = \Bigl[ f({x}, t) - g(t)^2\, \nabla_{{x}} \log p_t({x}) \Bigr] \diff t + g(t) \diff \hat{w}, \quad {x}(T) \sim p_T({x}),
    \label{eq:reverse-sde}
\end{equation}
where $\hat{w}$ is a reverse-time Wiener process, and $\nabla_{{x}} \log p_t({x})$ is called the score (or score function). The score $\nabla_{{x}} \log p_t({x})$ is the vector field of $x$ pointing to the directions in which the probability density function has the largest growth rate~\cite{song2019generative}. 
Simulating Eq.~\eqref{eq:reverse-sde} in time allows us to sample new data from noise. 

Earlier works such as the score-based generative models (SGMs)~\cite{song2019generative} often learn the score using \emph{score matching}~\cite{hyvarinen2005estimation}. However, score matching is computationally costly and only works for discrete times. \citet{song2020score} then propose a continuous-time version that optimises the following:
\begin{equation}
    \mathbb{E}_{t, x(0), x(t)} \Bigl[ \| s_\theta(x(t), t) - \nabla_{x(t)} \log p_{t} (x(t) \cond x(0)) \|^2 \Bigr],
    \label{eq:sde-score-matching}
\end{equation}
where $t$ is uniformly sampled over $[0, T]$, $x(0) \sim p_0(x)$, $x(t) \sim p_t(x(t) \mid x(0))$, and $s_\theta(x(t), t)$ represents the score prediction network. This objective ensures that the optimal score network, denoted $s_\theta^\ast (x(t), t)$, from Eq.~\eqref{eq:sde-score-matching} satisfies $s_\theta^\ast (x(t), t) = \nabla_x \log p_t(x)$ almost surely~\cite{vincent2011connection,song2020score}. 
{\small
\begin{tcolorbox}[title={\textbf{Block 6:} Extra reading - Score-based generative models (SGMs)}]\label{block6}
At the core of SGMs is the score (or score function) which can be connected to other diffusion-style approaches like DDPM, SDEs/ODES, and their combinations. To begin with, let's recall the energy-based models (EBMs)~\cite{lecun2006tutorial,song2021train,gustafsson2020train} that directly model the probability density function with learnable parameter $\theta$ as:
\begin{equation}
    p_\theta(x) = \frac{\expp^{-f_\theta(x)}}{Z_\theta}, \nonumber
\end{equation}
where $Z_\theta = \int \expp^{-f_\theta(x)} \diff x$ is a normalizing constant such that $\int p_\theta(x)\diff x = 1$, and $f_\theta(x)$ is an arbitrary parameterized function often called the unnormalized probabilistic model or energy-based model~\cite{lecun2006tutorial}.
Note that we cannot learn the model by directly maximizing its log-likelihood since $Z_\theta$ is intractable. Instead, one way to avoid calculating $Z_\theta$ is to learn the \textit{score function} $\nabla_x \log p(x)$ of the distribution $p(x)$. Taking the log derivative of both sides of the above equation gives that
\begin{equation}
    \nabla_x \log p_\theta(x) = \nabla_x \log \left(\frac{\expp^{-f_\theta(x)}}{Z_\theta} \right) = \nabla_x \log \expp^{-f_\theta(x)} - \bcancel{\nabla_x \log Z_\theta} = -\nabla_x f_\theta(x). \nonumber
\end{equation}
The term $-\nabla_x f_\theta(x)$ can be approximated with a neural network $s_\theta (x)$ which is called the score-based model representing the parameterized score function. Then we learn it by minimizing the Fisher divergence~\cite{fisher1922mathematical} between $s_\theta(x)$ and the ground truth score: $\mathbb{E}_{p(x)} \Bigl[ \| s_\theta(x) - \nabla_x \log p(x) \|^2 \Bigr].$ Intuitively, the score function defines a vector field over the entire data space that describes the direction that increases the likelihood of the data distribution. However, the ground truth score is always unknown and inaccessible. Then the \textit{score matching}~\cite{hyvarinen2005estimation} is proposed to optimize the score model without knowledge of the ground truth score. Specifically, it shows that the Fisher divergence is equivalent to the following objective:
\begin{equation}
    \mathbb{E}_{p(x)} \Bigl[ \text{tr}(\nabla_x s_\theta (x)) + \frac{1}{2}\| s_\theta(x) \|^2 \Bigr], \nonumber
\end{equation}
where $\nabla_x s_\theta (x)$ denotes the Jacobian of $s_\theta(x)$. This objective only contains $s_\theta(x)$ and thus is preferred for learning the score model. However, involving the Jacobian also means it is computationally costly when applied to high dimensional data. Then, the denoising score matching~\cite{vincent2011connection,song2019generative} technique is further proposed for efficient model optimization.

\paragraph{Denoising score matching}
This method first perturbs data with a pre-specified noise distribution $q_\sigma (\tilde{x} \cond x)$ and then learns the score of the perturbed data distribution $q_\sigma (\tilde{x}) \triangleq \int q_\sigma (\tilde{x} \cond x) p(x) \diff x$, which is equivalent to optimizing the following objective:
\begin{equation}
    \mathbb{E}_{q_\sigma (\tilde{x} \cond x) p(x)} \Bigl[ \| s_\theta(\tilde{x}) - \nabla_{\tilde{x}} \log q_\sigma (\tilde{x} \cond x) \|^2 \Bigr]. \nonumber
\end{equation}
Theoretically, the optimal score network (marked as $s_\theta^\ast (x)$) satisfies $s_\theta^\ast (x) = \nabla_x \log q_\sigma (x)$ almost surely~\cite{vincent2011connection}. However, we must make sure $q_\sigma (x) \approx p(x)$ i.e. use small noise perturbation such that $s_\theta^\ast (x) \approx \nabla_x \log p(x)$. Once trained, we can estimate the score and sample new data by simulating the Langevin Dynamics~\cite{grenander1994representations}.

\paragraph{Sampling with Langevin dynamics}
This is a well-known approach to sample data from noise with the score $\nabla_x \log p(x)$. At the key is an MCMC procedure that iterates the following~\cite{grenander1994representations}:
\begin{equation}
    x_{t} \leftarrow x_{t-1} + c \nabla_x \log p(x) + \sqrt{2c} \,\epsilon_t, \nonumber
\end{equation}
where $x_0 \sim \pi (x)$ is initialized from a prior distribution (such as a standard Gaussian) and $\epsilon_t \sim \mathcal{N}(0, I)$ is extra noise to ensure the diversity of the results. The coefficient $c$ is fixed as the step size that controls the speed of the sampling process. As $c \rightarrow 0$ and $t \rightarrow \infty$, the state $x_t$ converges to the data sampled from the true distribution $p(x)$. Note that the whole sampling process only requires the score $\nabla_x \log p(x)$ at each step, meaning that we can train a score network $s_\theta (x)$ to generate new samples by substituting $\nabla_x \log p(x)$ with $s_\theta (x)$.
\end{tcolorbox}
}%

\subsubsection{Interpreting DDPM with the variance preserving SDE}
Notably, extending DDPM to an infinite number of timesteps (i.e., continuous timesteps) leads to a special SDE which gives a more reliable interpretation of the diffusion process, and allows us to optimise the sampling with more efficient SDE/ODE solvers~\cite{song2020score,lu2022dpm}. Specifically, recall the DDPM perturbation kernel $q(x_t \mid x_{t-1})$ of Eq.~\eqref{eq:diffusion_forward_kernel} and write it in the form:
\begin{equation}
    x_i = \sqrt{1 - \beta_t} \, x_{i-1} + \sqrt{\beta_i} \, \epsilon_{i-1}, \quad \epsilon \sim \mathcal{N}(0, I) \;\; \text{and} \;\; i = 1, \cdots , N,
    \label{eq:discrete_diffusion_transition_kernel}
\end{equation}
where $i$ is the discrete timestep. Let us define an auxiliary set $\{ \bar{\beta}_i = N \beta_t \}_{i=1}^N$ and obtain
\begin{equation}
    x_i = \sqrt{1 - \frac{\bar{\beta}_i}{N}} \, x_{i-1} + \sqrt{\frac{\bar{\beta}_i}{N}} \, \epsilon_{i-1}.
    \label{eq:discrete_diffusion_transition_kernel_auxiliary}
\end{equation}
By further letting functions $\beta(\frac{i}{N}) \coloneqq \bar{\beta}_i$, $x(\frac{i}{N}) \coloneqq x_i$, $\epsilon(\frac{i}{N}) \coloneqq \epsilon_i$ (as a preparation to convert functions from discrete to continuous), we can rewrite Eq.~\eqref{eq:discrete_diffusion_transition_kernel_auxiliary} with the difference $\Delta t = \frac{1}{N}$ and time $t \in {0, \frac{1}{N}, \cdots , \frac{N-1}{N}}$  as follows:
\begin{align}
    x(t + \Delta t) 
    &= \sqrt{1 - \beta(t + \Delta t) \Delta t} \, x(t) + \sqrt{\beta (t + \Delta t)\Delta t} \, \epsilon(t) \\
    &\approx x(t) - \frac{1}{2}\beta(t)\Delta t \, x(t) + \sqrt{\beta (t + \Delta t)} \, \sqrt{\Delta t} \, \epsilon(t) \quad \quad \text{(Taylor series)} \\
    &\approx x(t) - \frac{1}{2}\beta(t)\Delta t \, x(t) + \sqrt{\beta(t)} \, \sqrt{\Delta t} \, \epsilon(t),
\end{align}
where the two approximate equalities hold when $\Delta t \rightarrow 0$. Then we convert $\Delta t$ to $\diff t$, $\sqrt{\Delta t} \, \epsilon(t)$ to $\diff w$ and obtain the following:
\begin{equation}
    \diff x = -\frac{1}{2} \beta(t) x \diff t + \sqrt{\beta(t)} \diff w,
    \label{eq:vp-sde}
\end{equation}
which is a typical mean-reverting SDE (also known as the Ornstein–Uhlenbeck process~\cite{gillespie1996exact}) that drifts towards a stationary distribution, i.e. a standard Gaussian in this case. \citet{song2020score} also name it the variance preserving (VP) SDE and further illustrate that DDPM's marginal distribution $q(x_t \mid x_0)$ in Eq.~\eqref{eq:diffusion_marginalize_kernel} is a solution to the VP-SDE. Therefore, we can use either the diffusion reverse process (Eq.~\eqref{eq:diffusion_sample}) or the reverse-time SDE (Eq.~\eqref{eq:reverse-sde}) to sample new data from noise with the same trained DDPM. In addition, the score $\nabla_{{x}} \log p_t({x})$ can be directly computed from the marginal distribution $q(x_t \mid x_0)$ in Eq.~\eqref{eq:diffusion_marginalize_kernel},
\begin{equation}
    \nabla_{x_t} \log p_t(x_t) = -\frac{x_t - \sqrt{\bar{\alpha}_t} x_0}{(1-\bar{\alpha}_t)} = - \frac{\epsilon_t}{\sqrt{1-\bar{\alpha}_t}},
    \label{eq:score_to_noise}
\end{equation}
where $\epsilon_t$ is from the reparameterization trick and can be approximated using the noise prediction network $\epsilon_\theta(x_t, t)$.  Eq.~\eqref{eq:score_to_noise} thus shows how we convert the diffusion model to an SDE (i.e., obtain the score $\nabla_{{x}} \log p_t({x})$ from noise $\epsilon_\theta(x_t, t)$). Then, numerous efficient SDEs/ODEs solvers can be used to optimise diffusion models, further bringing interpretability and faster sampling~\cite{song2020score}.

\subsection{Conditional Diffusion Models}
\label{subsec:guided_diffusion}

So far, we have learned how to sample data from different types of diffusion models. However, all the above methods only consider unconditional generation, which is insufficient for image restoration where we want to sample HQ images conditioned on degraded LQ images. Therefore, we present the \emph{conditional} diffusion model below.

Let us keep the diffusion process $q(x_{1:T} \mid x_0)$ of Eq.~\eqref{eq:diffusion_forward} unchanged and reconstruct the reverse process in Eq.~\eqref{eq:diffusion_reverse} with a condition $y$, i.e. $p_\theta(x_{0:T} \mid y) = p(x_T \mid y) \prod_{t=1}^T p_{\theta}(x_{t-1} \mid x_t, \, y)$. The conditional reverse kernel can then be modeled as
\begin{align}
    p_{\theta, \phi}(x_{t-1} \mid x_t, \, y) = Z\cdot p_{\theta}(x_{t-1} \mid x_{t}) \, p_{\phi}(y \mid x_{t-1}),
    \label{eq:condition_reverse_kernel}
\end{align}
where $p_\phi(y \mid x)$ is an additional network that predicts $y$ from $x$, and $Z = p_{\phi}(y \mid x_{t})^{-1}$ can be treated as a constant since it does not depend on $x_{t-1}$.
{\small
\begin{tcolorbox}[title={\textbf{Block 7:} Complete derivation for the conditional reverse kernel of Eq.~\eqref{eq:condition_reverse_kernel}}]\label{block7}
\begin{proof}
We can first derive a fact that $p_{\phi}(y \mid x_{t-1}, x_{t})$ doesn't depend on $x_t$:
\begin{equation}
\begin{aligned}
    p_{\phi}(y \mid x_{t-1}, x_{t}) 
    &= p_{\theta}(x_{t} \mid x_{t-1}, y) \frac{p_\phi(y \mid x_{t-1})}{p_\theta(x_t \mid x_{t-1})} \\
    &= p_{\theta}(x_{t} \mid x_{t-1}) \frac{p_\phi(y \mid x_{t-1})}{p_\theta(x_t \mid x_{t-1})} \\
    &= p_\phi(y \mid x_{t-1}), \nonumber
\end{aligned}
\end{equation}
which gives the following conditional reverse distribution:
\begin{equation}
    \begin{aligned}
    p_{\theta}(x_{t-1} \mid x_t, \, y)
    &= \frac{p_{\theta}(x_{t-1}, \, x_t, \, y)}{p_{\theta}(x_t, \, y)} \\
    &= \frac{p_{\phi}(y \mid x_{t-1}, x_{t}) \, p_{\theta}(x_{t-1} \mid x_{t}) \, p(x_t)}{p_{\phi}(y \mid x_{t}) \, p(x_t)} \\
    &= \frac{p_{\phi}(y \mid x_{t-1}, x_{t}) \, p_{\theta}(x_{t-1} \mid x_{t})}{p_{\phi}(y \mid x_{t})} \\
    &= \frac{p_{\phi}(y \mid x_{t-1}) \, p_{\theta}(x_{t-1} \mid x_{t})}{p_{\phi}(y \mid x_{t})}. \nonumber
    \end{aligned}
\end{equation}
Note that $p_{\phi}(y \mid x_{t})$ does not depend on $x_{t-1}$ thus it can be treated as a constant $Z^{-1}$. Therefore, the conditional reverse kernel can be written as
\begin{equation}
    p_{\theta, \phi}(x_{t-1} \mid x_t, \, y) = Z p_{\theta}(x_{t-1} \mid x_{t}) \, p_{\phi}(y \mid x_{t-1}), \nonumber
\end{equation}
which then completes the proof.
\end{proof}
\end{tcolorbox}
}%
This equation yields an adjusted mean for the posterior distribution of Eq.~\eqref{eq:learnable_diffusion_reverse_kernel}, given by~\cite{dhariwal2021diffusion}:
\begin{equation}
    \hat{\mu}_\theta(x_t, t, y) = \mu_\theta(x_t, t) + \eta \cdot \tilde{\beta}_t \nabla_{x_t} \log p_\phi(y \mid x_t),
    \label{eq:condition_guidence_mean}
\end{equation}
where $\eta$ is the gradient scale (also called the guidance scale). Moreover, recall that the score can be approximated using the noise prediction network: $\nabla_{x_t} \log p_t(x_t) \approx - \frac{1}{\sqrt{1-\bar{\alpha}_t}} \epsilon_\theta(x_t, t)$ from Eq.~\eqref{eq:score_to_noise}, which further gives the score of the joint distribution $p_t(x_t, y)$:
\begin{align}
    \nabla_{x_t} \log p_t(x_t, y) 
    &= \nabla_{x_t} \log p_t(x_t) + \nabla_{x_t} \log p_t(y \cond x_t) \\
    &\approx - \frac{1}{\sqrt{1-\bar{\alpha}_t}} \epsilon_\theta(x_t, t) + \nabla_{x_t} \log p_\phi(y \cond x_t) \\
    &= - \frac{1}{\sqrt{1-\bar{\alpha}_t}} \bigl(\epsilon_\theta(x_t, t) - \sqrt{1-\bar{\alpha}_t} \nabla_{x_t} \log p_\phi(y \cond x_t) \bigr),
\end{align}
which provides a conditional noise predictor $\hat{\epsilon}_\theta$ with the following form~\cite{dhariwal2021diffusion}: 
\begin{equation}
    \hat{\epsilon}_\theta(x_t, t, y) = \epsilon_\theta(x_t, t) - \eta \cdot \sqrt{1-\bar{\alpha}_t} \nabla_{x_t} \log p_\phi(y \cond x_t).
    \label{eq:guided_noise_predictor}
\end{equation}
The conditional sampling is performed as a regular DDPM by substituting the new noise predictor $\hat{\epsilon}_\theta(x_t, t, y)$ into the posterior mean of Eq.~\eqref{eq:diffusion_learnable_posterior_mean}. The gradient scale $\eta$ controls the performance trade-off between image quality and fidelity, i.e. lower guidance scale produces photo-realistic results, and higher guidance scale yields better consistency with the condition.
{\small
\begin{tcolorbox}[title={\textbf{Block 8:} Complete derivation for the adjusted mean of Eq.~\eqref{eq:condition_guidence_mean}}]\label{block8}
\begin{proof}
    For notation simplicity, we set $p_\theta(x_t \mid x_{t+1}) = \mathcal{N}(\mu, \Sigma)$ and then having that 
    \begin{equation}
        \log p_\theta(x_t \mid x_{t+1}) = -\frac{1}{2} (x_t - \mu)^\mathsf{T} \Sigma^{-1}(x_t - \mu) + C, \nonumber
    \end{equation}
    where $C$ is a constant. And the term $\log p_\phi(y \mid x_t)$ can be approximated using Taylor expansion around $x_t = \mu$ as the following:
    \begin{equation}
    \begin{aligned}
        \log p_\phi(y \mid x_t) 
        &\approx \log p_\phi(y \mid x_t) |_{x_t=\mu} + (x_t - \mu) \nabla_{x_t} \log p_\phi(y \mid x_t)|_{x_t=\mu} \\
        &= (x_t - \mu)g + C_1, \nonumber
    \end{aligned}
    \end{equation}
    
    where $g=\nabla_{x_t} \log p_\phi(y \mid x_t)|_{x_t=\mu}$ and $C_1$ is a constant. Then we can compute
    \begin{equation}
    \begin{aligned}
        \log \bigl(p_{\theta}(x_t \mid x_{t+1}) \, p_{\phi}(y \mid x_{t}) \bigr)
        &\approx -\frac{1}{2} (x_t - \mu)^\mathsf{T} \Sigma^{-1}(x_t - \mu) + (x_t - \mu)g + C_2 \\
        &= -\frac{1}{2} (x_t - \mu - \Sigma g)^\mathsf{T} \Sigma^{-1}(x_t - \mu - \Sigma g) + C_3 \\
        &= \log p(z) + C_4, \; z \sim \mathcal{N}(\mu + \Sigma g, \Sigma), \nonumber
    \end{aligned}
    \end{equation}
    where $C_2, C_3, C_4$ are constants and $C_4$ can be ignored as the normalizing coefficient $Z$ in Eq.~\eqref{eq:condition_reverse_kernel}. We assume that the conditional reverse kernel is also a Gaussian i.e. $p(x_t \mid x_{t+1}, \, y) \sim \mathcal{N}(\hat{\mu}, \Sigma)$. By substituting parameters with the real transition kernel $p_{\theta}(x_{t-1} \mid x_t) = \mathcal{N}(x_{t-1}; \mu_\theta(x_t, t), \tilde{\beta}_t I)$ and adding the gradient scale $\eta$ to $g$, we obtain 
    \begin{equation}
        \hat{\mu}_\theta(x_t, t, y) = \mu_\theta(x_t, t) + \eta \cdot \tilde{\beta}_t \nabla_{x_t} \log p_\phi(y \mid x_t), \nonumber
    \end{equation}
    which is the adjusted mean and thus completes the proof.
\end{proof}
\end{tcolorbox}
}%
\paragraph{Conditional SDE}
Similar to guided diffusion, we can also change the score function to control the reverse-time SDE conditioned on the variable $y$, i.e. by replacing $\nabla_{{x}} \log p_t({x})$ with $\nabla_x \log p_t(x \cond y)$ in Eq.~\eqref{eq:reverse-sde}. Since $p_t(x \cond y) \propto p_t(x) \, p_t(y \cond x)$, the conditional score can be decomposed as 
\begin{equation}
    \nabla_x \log p_t(x \cond y) = \nabla_x \log p_t(x) + \nabla_x \log p_t(y \cond x),
    \label{eq:condition_score}
\end{equation}
which means that we can simulate the following reverse-time SDE for conditional generation:
\begin{equation}
    \diff {x} = \Bigl[ f({x}, t) - g(t)^2\, \bigl(\nabla_{{x}} \log p_t({x}) + \nabla_x \log p_t(y \cond x) \bigr) \Bigr] \diff t + g(t) \diff \hat{w},
    \label{eq:condition_reverse-sde}
\end{equation}
where ${x}(T) \sim p_T(x \cond y)$. \citet{song2020score} show that we can use a separate network to learn $p_t(y \cond x)$ (e.g., a time-dependent classifier if $y$ represents class labels), or estimate its log gradient $\nabla_x \log p_t(y \cond x)$ directly with heuristics and domain knowledge. 

With these conditional diffusion models, we can sample images with specified labels (such as dog and cat) or, as the main topic of this paper, recover clean HQ images from corrupted LQ inputs.\looseness=-1

\section{Diffusion Models for Image Restoration}\label{sec:dm4ir}

Diffusion-based image restoration (IR) can be considered a special case of conditional diffusion models with image conditioning. We first introduce the concept of image degradation, which is a process that transforms a high-quality (HQ) image $x$ to a low-quality (LQ) image $y$ characterized by undesired corruptions. The general image degradation process can be modelled as follows:
\begin{equation}
    y = A(x) + n,
    \label{eq:ir_problem}
\end{equation}
where $A$ denotes the degradation function and $n$ is additive noise. As the examples show in Figure~\ref{fig:ir-tasks}, degradation can manifest in various forms such as noise, blur, rain, haze, etc. IR then aims to reverse this process to obtain a clean HQ image from the corrupted LQ counterpart $y$. 

IR is further decomposed into two distinct settings, \emph{blind and non-blind IR}, depending on whether or not the degradation parameters $A$ and $n$ of Eq.~\eqref{eq:ir_problem} are known. \emph{Blind IR} is the most general setting, in which no explicit knowledge of the degradation process is assumed. Blind IR methods instead utilize datasets of paired LQ-HQ images for supervised training of models. \emph{Non-blind IR} methods, in contrast, assume access to $A$ and $n$. This is an unrealistic assumption for many important real-world IR tasks, and thus limits non-blind methods to a subset of specific IR tasks such as bicubic downsampling, Gaussian blurring, colorization, or inpainting with a fixed mask. In the following, we first describe the most straightforward diffusion-based approach for general \emph{blind} IR tasks in Sec.~\ref{subsec:cddm}. Representative \emph{non-blind} diffusion-based approaches are then covered in Sec.~\ref{subsec:training_free}. Lastly, Sec.~\ref{sec:diffusion_towards_lq} covers more recent methods for general \emph{blind} IR.

\subsection{Conditional Direct Diffusion Model}\label{subsec:cddm}

\begin{figure}[t]
    \centering
    \includegraphics[width=1.\linewidth]{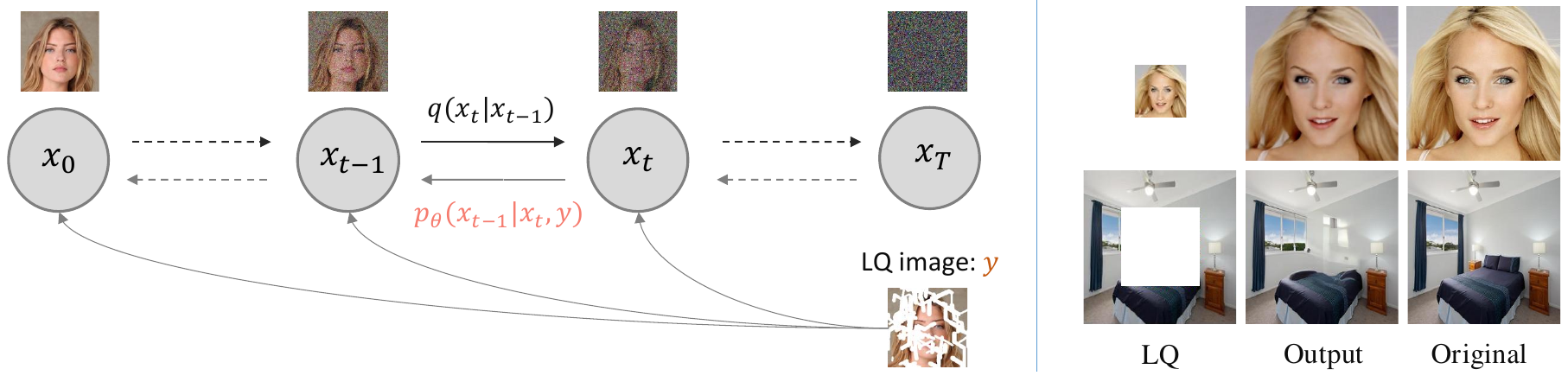}\vspace{-1.0mm}
    \caption{\textit{Left:} Overview of the conditional direct diffusion model (CDDM) on the face inpainting case. The only change compared to DDPM (Figure~\ref{fig:ddpm}) is the reverse transition model $p_{\theta}(x_{t-1} \mid x_t, \, y)$, which involves the LQ image $y$ in sampling to generate the corresponding HQ image. \textit{Right:} Two image restoration examples (image super-resolution and inpainting) performed under the CDDM framework. These results look realistic but are not consistent with the original image.}\vspace{-1.0mm}
    \label{fig:direct-diffusion}
\end{figure}

The most straightforward approach for applying DMs to general IR tasks is to use the conditional diffusion model (CDM) with image guidance from Sec.~\ref{sec:gmdm}~\ref{subsec:guided_diffusion}. In the IR context, the term $p_\phi(y \mid x)$ in Eq.~\eqref{eq:guided_noise_predictor} represents the image degradation model which can be either a fixed operator with known parameters or a learnable neural network, depending on the task. It is also noted that strong guidance (large $\eta$ in Eq.~\eqref{eq:guided_noise_predictor}) leads to good fidelity but visually lower-quality results (e.g., over-smooth images), while weak guidance (small $\eta$) has the opposite effect~\cite{dhariwal2021diffusion}. Now, let us consider the extreme case: \emph{how about decreasing $\eta$ to zero, i.e. no guidance?} A simple observation from Eq.~\eqref{eq:guided_noise_predictor} is that with $\eta = 0$, the conditional noise predictor learns the unconditional noise predictor directly: $\hat{\epsilon}_\theta(x_t, t, y) = \epsilon_\theta(x_t, t)$, and the objective for diffusion-based IR is given by
\begin{equation}
    L_{\mathrm{cddm}} = \mathbb{E}_{x_0, y, t, \epsilon} \Bigl[\| \epsilon_t - \hat{\epsilon}_\theta(\sqrt{\bar{\alpha}_t}x_0 + \sqrt{1-\bar{\alpha}_t}\cdot \epsilon, t, y) \|^2 \Bigr].
    \label{eq:direct_condition_diffusion_loss}
\end{equation}
We name this the conditional direct diffusion model (CDDM), which essentially follows the same training and sampling procedure as DDPM, except for the condition $y$ in noise prediction as shown in Figure~\ref{fig:direct-diffusion}. As a result, the generated image can be of very high visual quality (it looks realistic), but often has limited consistency with the original HQ image~\cite{saharia2022image,rombach2022high}, as can be observed for the examples in the right of Figure~\ref{fig:direct-diffusion}. Fortunately, some IR tasks, such as image super-resolution, colorization and inpainting, are highly ill-posed and can tolerate diverse predictions. CDDM can then be effectively trained on these tasks as a supervised approach for photo-realistic image restoration. 

One typical method is SR3~\cite{saharia2022image}, which employs CDDM with a few modifications for image super-resolution. To condition the model on the LQ image $y$, SR3 up-samples $y$ to the target resolution so that $y$ can be concatenated with the intermediate state $x_t$ along the channel dimension. 
Subsequently, Palette~\cite{saharia2022palette} extends SR3 to general IR tasks including colorization, inpainting, uncropping, and JPEG restoration. Various other works~\citep{whang2022deblurring,ozdenizci2023restoring,jiang2023low} also take the same `direct diffusion' strategy but adopt different restoration pipelines and additional networks for task-specific model learning. More recently, \citet{wang2024exploiting} propose StableSR, which further adapts a large-scale pretrained diffusion model (Stable Diffusion~\cite{rombach2022high}) for image restoration, by tweaking the noise predictor with image conditioning in the same way as for CDDM.

\subsection{Training-free Conditional Diffusion Models}\label{subsec:training_free}

The key to the success of CDDM in image restoration lies in learning the conditional noise predictor $\hat{\epsilon}_\theta (x_t, t, y)$ by optimising Eq.~\eqref{eq:direct_condition_diffusion_loss} on a dataset of paired LQ-HQ images. Unfortunately, this means that $\hat{\epsilon}_\theta (x_t, t, y)$ needs to be re-trained to handle tasks which are not included in the current training data, even in the non-blind setting where the degradation parameters $A$ and $n$ in Eq.~\eqref{eq:ir_problem} are known. For non-blind IR, a \emph{training-free} approach can instead be derived by directly incorporating the degradation function into a pretrained unconditional diffusion model, such as a DDPM.

With known degradation parameters, the term $p(y \mid x)$ also becomes accessible: $p(y \mid x) = \mathcal{N}(A(x), \sigma^2_n I)$, if the noise $n$ is Gaussian. Traditional IR approaches often solve this problem using maximum a posteriori (MAP) estimation~\cite{banham1997digital}, as follows:
\begin{equation}
    \hat{x} = \arg \min_x \frac{1}{2\sigma^2_n} \| y - A(x) \|^2 + \lambda \mathcal{P}(x),
    \label{eq:map}
\end{equation}
where $\mathcal{P}(x)$ is a prior term empirically chosen to characterize the prior knowledge of $x$. Then, a natural idea is to incorporate a pretrained unconditional DDPM into $\mathcal{P}(x)$ as a powerful learned image prior.
Specifically, recall the conditional score of Eq.~\eqref{eq:condition_score} in the form:
\begin{equation}
    \nabla_{x_t} \log p_t(x_t \cond y) = \nabla_{x_t} \log p_t(x_t) + \nabla_{x_t} \log p_t(y \cond x_t),
    \label{eq:discrete_condition_score}
\end{equation}
where $x_t$ matches the diffusion state in DDPM, and the unconditional score $\nabla_{x_t} \log p_t(x_t)$ can be obtained from Eq.~\eqref{eq:score_to_noise} and approximated with DDPM's noise predictor, as $\nabla_{x_t} \log p_t(x_t) \approx s_\theta(x_t, t) = - \frac{\epsilon_\theta(x_t, t)}{\sqrt{1-\bar{\alpha}_t}}$. Computing $p_t(y \mid x_t)$ in \eqref{eq:discrete_condition_score} is however difficult since there is no obvious relationship between $y$ and state $x_t$. Fortunately, with Gaussian noise $n \sim \mathcal{N}(0, \sigma^2_n I)$, \citet{chung2022diffusion} propose an approximation for $\nabla_{x_t} \log p_t(y \cond x_t)$ at each timestep $t$:
\begin{equation}
    \nabla_{x_t} \log p_t(y \cond x_t) \approx \nabla_{x_t} \log p_t(y \cond \hat{x}_0), \quad \mathrm{where} \quad \hat{x}_0 = \frac{1}{\sqrt{\bar{\alpha}_t}} (x_t + (1 - \bar{\alpha}_t)s_\theta(x_t, t)).
    \label{eq:cond_score_approx}
\end{equation}
This can be obtained via Tweedie's formula~\citep{efron2011tweedie,chung2022diffusion,song2023pseudoinverse,boys2023tweedie}. The approximation above is motivated by a Dirac approximation $\nabla_{x_t} \log p(y \mid x_t) = \nabla_{x_t} \log  \int p(y \mid x_0) \, p(x_0 \mid x_t) d x_0 \approx \nabla_{x_t} \log p(y \mid \hat{x}_0(x_t))$, where $\hat{x}_0(x_t)$ is any Monte Carlo sample of $p(x_0 \mid x_t)$. However, the approximation usually exhibits high variance since it uses a single Monte Carlo sample, while drawing more samples incurs more computations. 
Moreover, it is worth noting that $p_t(y \mid \hat{x}_0)$ is a tractable Gaussian: $p_t(y \mid \hat{x}_0) = \mathcal{N}(A(\hat{x}_0), \sigma^2_n I)$. Computing $\nabla_{x_t} \log p_t(y \cond \hat{x}_0)$ and substituting it for $\nabla_{x_t} \log p_t(y \cond x_t)$ in Eq.~\eqref{eq:discrete_condition_score} thus gives the following:
\begin{equation}
    \nabla_{x_t} \log p_t(x_t \cond y) \approx s_\theta(x_t, t) - \frac{1}{2 \sigma^2_n} \nabla_{x_t} \| y - A(\hat{x}_0) \|^2.
    \label{eq:cond_score_approx2}
\end{equation}
We can then incorporate this approximation Eq.~\eqref{eq:cond_score_approx2} into the sampling of a pretrained DDPM,
\begin{align}
    x_{t-1} 
    &= \frac{1}{\sqrt{\bar{\alpha}_t}}(x_t + (1 - \alpha_t) \nabla_{x_t} \log p_t(x_t \cond y)) + \sqrt{\tilde{\beta}_t}\epsilon \\
    &\approx \frac{1}{\sqrt{\bar{\alpha}_t}}(x_t + (1 - \alpha_t) [s_\theta(x_t, t) - \frac{1}{2 \sigma^2_n} \nabla_{x_t} \| y - A(\hat{x}_0) \|^2]) + \sqrt{\tilde{\beta}_t}\epsilon \\
    &= \underbrace{- \frac{1 - \alpha_t}{2 \sigma^2_n \sqrt{\bar{\alpha}_t}} \nabla_{x_t} \| y - A(\hat{x}_0) \|^2}_{\text{data consistency term}} + \underbrace{\frac{1}{\sqrt{\bar{\alpha}_t}}(x_t + (1 - \alpha_t)s_\theta(x_t, t)) + \sqrt{\tilde{\beta}_t}\epsilon}_{\text{diffusion term}},
    \label{eq:dps_detail}
\end{align}
where the first line is derived from Eq.~\eqref{eq:diffusion_sample} and Eq.~\eqref{eq:score_to_noise} with additional condition $y$. Note that the diffusion term is actually an unconditional sampling step in DDPM, where $s_\theta$ is obtained from Eq.~\eqref{eq:score_to_noise} as $s_\theta(x_t, t) = - \frac{\epsilon_\theta(x_t, t)}{\sqrt{1-\bar{\alpha}_t}}$. By letting $\rho=\frac{1 - \alpha_t}{2 \sigma^2_n \sqrt{\bar{\alpha}_t}}$ represent the step size of the data consistency term and simplifying the diffusion term, we then finally have:
\begin{equation}
    x_{t-1} = - \rho \nabla_{x_t} \| y - A(\hat{x}_0) \|^2 + \mu_\theta(x_t, t) + \sqrt{\tilde{\beta}_t} \, \epsilon,
    \label{eq:dps}
\end{equation}
where $\mu_\theta$ and $\tilde{\beta}$ are the posterior mean and variance of Eq.~\eqref{eq:learnable_diffusion_reverse_kernel}, respectively. This approach is called the diffusion posterior sampling (DPS)~\cite{chung2022diffusion}. Note that Eq.~\eqref{eq:dps} is conceptually similar to the MAP estimation of Eq.~\eqref{eq:map}, with $\nabla_{x_t} \| y - A(\hat{x}_0) \|^2$ as the data consistency term and $\mu_\theta(x_t, t) + \sqrt{\tilde{\beta}_t} \, \epsilon$ being a diffusion-based image prior. When the degradation parameters of Eq.~\eqref{eq:ir_problem} are known, DPS thus utilizes this knowledge to guide the sampling process of a pretrained DDPM, encouraging generated images to be consistent with the LQ input $y$.

\begin{figure}[t]
    \centering
    \includegraphics[width=1.\linewidth]{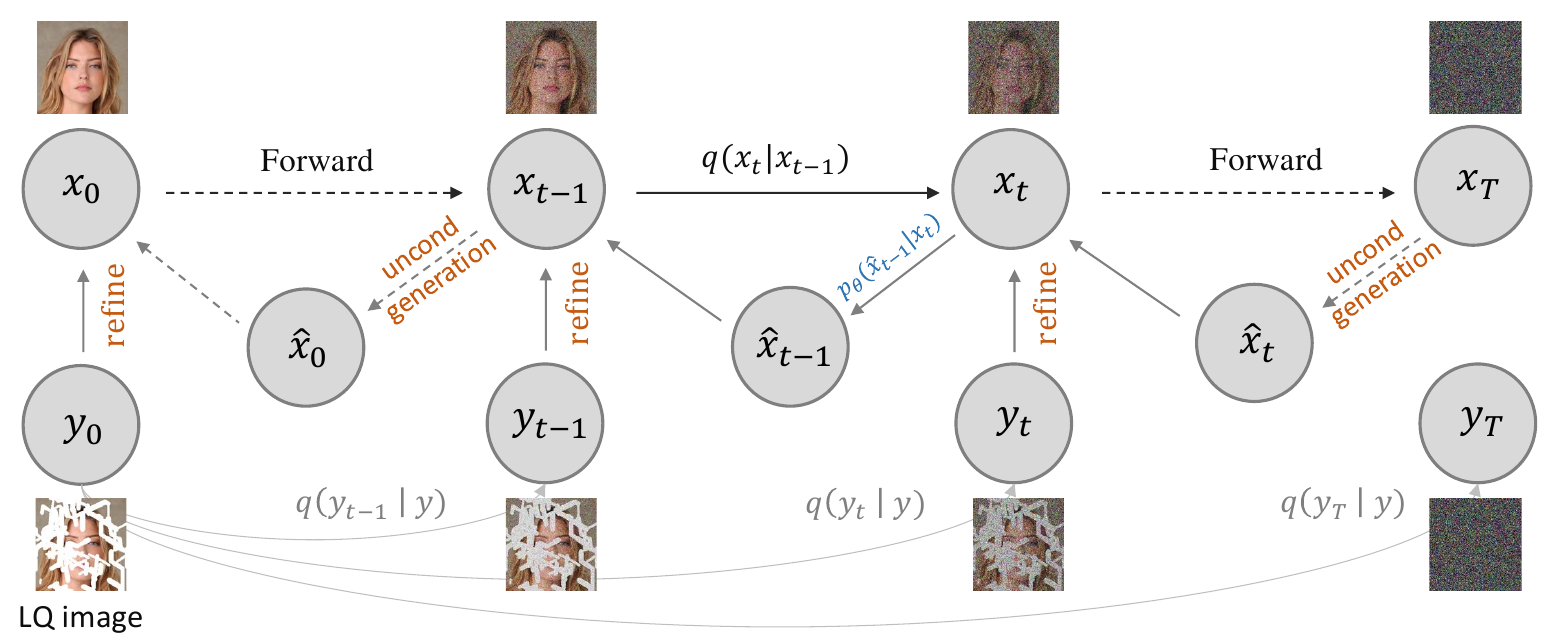}
    \caption{Overview of the projection-based CDM. There are two paths for the HQ image $x$ and LQ image $y$, generated from the same diffusion model. At each reverse step $t$, the sampling first leverages the pretrained DDPM for unconditional generation, i.e. $p_\theta (\hat{x}_t \mid x_{t+1})$, and then refines $\hat{x}_t$ to $x_t$ with functions $H$ and $b$ as $x_t = H(\hat{x}_t) + b(y_t)$, where $y_t$ is obtained by applying the forward marginal transition Eq.~\eqref{eq:diffusion_marginalize_kernel} on the LQ image as $y_t \sim q(y_t \mid y)$.}
    \label{fig:train-free-cdm}
\end{figure}

DPS does however rely on the approximation in Eq.~\eqref{eq:cond_score_approx}, for which the approximation error approaches 0 only when the noise $n$ of $y$ has a high variance: $\sigma_n \rightarrow \infty$.
For the case where the LQ image is noiseless, $y = A(x)$, we would prefer to introduce the approach from Figure~\ref{fig:train-free-cdm} in which the unconditional generated state $\hat{x}_t$ is refined using degradation $A$ and the LQ image $y$. More specifically, since now the term $\nabla_{x_t} \log p_t(y \cond x_t)$ is unattainable (or non-approximable), we instead apply the same diffusion process to $y$ and obtain $p(y_t \mid y)=\mathcal{N}(y_t; \sqrt{\bar{\alpha}_t} y, (1-\bar{\alpha}_t)I)$, where $y_t$ corresponds to the degraded version of the state $x_t$. Then, we impose the data consistency by projecting the unconditional state onto a conditional path as follows:
\begin{equation}
    x_t = H(\hat{x}_t) + b(y_t), \quad \mathrm{where} \quad \hat{x}_t = \mu_\theta(x_t, t) + \sqrt{\tilde{\beta}_t} \, \epsilon,
    \label{eq:projection}
\end{equation}
where $H$ and $b$ are functions derived from the known degradation $A$. For computational efficiency, the two functions are typically assumed to be linear and tailored to specific tasks. This projection-based method is also called the iterative latent variable refinement~\citep{choi2021ilvr,chung2022come,chung2022improving}. In addition, for linear degradation problems, we can further decompose $A$ into partitions and then combine them with the LQ image $y$ to refine the intermediate state $x_t$ in the reverse diffusion process~\cite{kawar2022denoising,wang2022zero}. This is similar to the projection-based approach but can be more computationally efficient since there is no need to compute $y_t$ for each reverse step.

Recently, another class of approaches for training-free conditional diffusion models have emerged which are based on Feynman--Kac models and sequential Monte Carlo (SMC) samplers~\citep{Wu2023smc, cardoso2024monte, janati2024divide}. 
At the core, they wrap the approximations for $p_t(y \mid x_t)$ in the proposals of an SMC sampler, so that the marginal distributions of the sampler anneals to the target conditional one. 
This approach is statistically exact, regardless of the approximations in $p_t(y \mid x_t)$, in the sense that as the number of particles used in the SMC sampler goes to infinity, the resulting population converges in distribution to the target. 
As such, this type of method improves significantly over DPS~\cite{chung2022diffusion} in terms of statistical errors. 
However, it comes at the cost of storing a population of particles which does not scale well in memory in the problem dimension, and the efficiencies of their proposals. 
To improve the sampler, \citet{cardoso2024monte} consider linear Gaussian likelihood models and propose efficient proposals based on an inpainting problem, while ~\citet{janati2024divide} develop a divide-and-conquer construction to set intermediate target distributions. 
Note that although~\citet{dou2024diffusion, corenflos2024conditioning} also use SMC samplers, they target different Feynman--Kac models. 
Moreover, the methods in~\cite{corenflos2024conditioning} are training-free only for special problems (e.g., inpainting).

\subsection{Diffusion Process towards Degraded Images}\label{sec:diffusion_towards_lq}

In previous sections, we have presented several diffusion-based IR methods, both for the blind and non-blind setting. However, these methods all generate images starting from Gaussian noise, which intuitively should be inefficient for IR tasks, given that input LQ images are closely related to the corresponding HQ images. That is, it should be easier to translate directly from LQ to HQ image, rather than from noise to HQ image. To address this problem, for general blind IR tasks, \citet{luo2023image} propose the IR-SDE that models image degradation with a mean-reverting SDE:
\begin{equation}
    dx = \theta_t \, (\mu - x) d t + \sigma_t d w, 
    \label{eq:ou}
\end{equation}
where $\mu$ is the state mean the SDE drifts to. $\theta_t$ and $\sigma_t$ are predefined coefficients that control the speed of the mean-reversion and the stochastic volatility, respectively. It is noted that the VP-SDE~\cite{song2020score} is a special case of Eq.~\eqref{eq:ou} where $\mu$ is set to 0. Moreover, the SDE in Eq.~\eqref{eq:ou} is proven to be tractable when the coefficients satisfy $\sigma_t^2 \, / \, \theta_t = 2 \, \lambda^2$ for all timesteps~\cite{luo2023image}, where $\lambda^2$ is the stationary Gaussian variance. Similar to DDPM, we can obtain the marginal transition kernel $p_t(x)$, which is a Gaussian given by
\begin{equation}
    p_t(x_t \mid x_0) = \mathcal{N}\Bigl(x_t \mid \mu + (x_0 - \mu) \, \expp^{-\bar{\theta}_{t}}, \lambda^2 \, (1 - \expp^{-2 \, \bar{\theta}_{t}})\Bigr),
\label{eq:irsde_solution}
\end{equation}
where $\bar{\theta}_{t} = \int^t_0 \theta_z \diff z$. As $t \to \infty$, the terminal distribution converges to a stationary Gaussian with mean $\mu$ and variance $\lambda^2$. By setting the HQ image as the initial state $x_0$ and the LQ image as the terminal state mean $\mu$, this SDE iteratively transforms the HQ image into the LQ image with additional noise (where the noise level is fixed to $\lambda$). Then, we can restore the HQ image based on the reverse-time process of Eq.~\eqref{eq:ou} as follows:
\begin{equation}
    dx = \big[ \theta_t \, (\mu - x) - \sigma_t^2 \, \nabla_{x} \log p_t(x) \big] d t + \sigma_t d \hat{w}.
    \label{eq:reverse-irsde}
\end{equation}
Notably, the score function $\nabla_{x} \log p_t(x)$ is tractable when conditioning on the known $x_0$ in training, as 
\begin{equation}
    \nabla_{x}\log p_t(x \cond x_0) = - \frac{x_t - m_t}{v_t}
\end{equation}
where $m_t$ and $v_t$ are the mean and variance of Eq.~\eqref{eq:irsde_solution}, respectively. Learning this score with a neural network is similar to denoising score matching~\cite{vincent2011connection} but the target score is directly computed from the training distributions.

\begin{figure}[t]
    \centering
    \includegraphics[width=.8\linewidth]{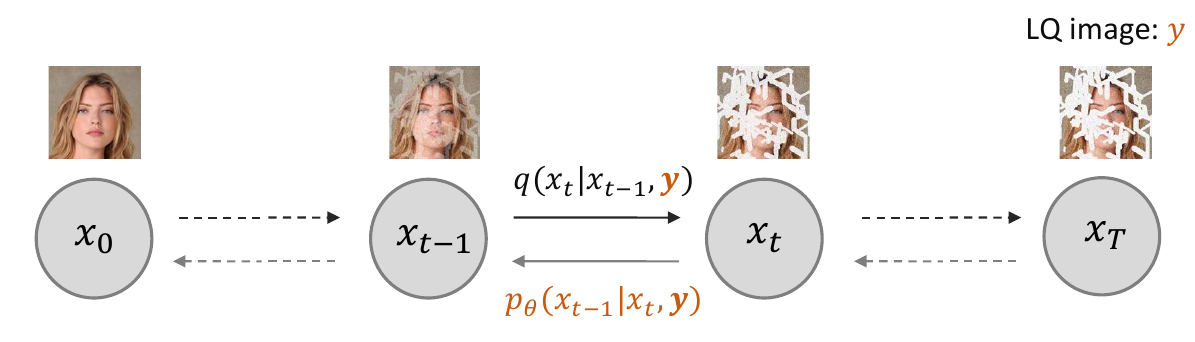}
    \caption{Overview of the approach that performs diffusion towards degraded images. Here, the LQ image $y$ is involved in both the forward and backward processes. Moreover, the terminal state $x_T$ is often a (noisy) LQ image rather than the Gaussian noise.}
    \label{fig:p2p-diffusion}
\end{figure}

However, IR-SDE still needs to add noise to the LQ image as a terminal state $x_T$. For fixed point-to-point mapping with a diffusion process, we further introduce the diffusion bridge~\cite{li2023bbdm} which can naturally transfer complex data distributions to reference distributions, i.e. directly from HQ to LQ images, without adding noise. More specifically, given a diffusion process defined by a forward SDE as in Eq.~\eqref{eq:forward-sde}, \citet{rogers2000diffusions} show that we can force the SDE to drift from HQ image $x$ to a particular condition (the degraded image $y$) via Doob's $h$-transform~\cite{doob1984classical}:
\begin{equation}
	\diff {x} = f(x, t) \diff t + {g(t)}^2 h(x_t,t,y,T) + g(t)\diff w, 
	\label{eq:diffusion-bridge}
\end{equation}
where $h(x_t,t,y,T)=\nabla_{x_t} \log p(x_T \cond x_t) \cond_{x_T=y}$ is the gradient of the log transition kernel from $t$ to $T$, derived from the original SDE. By setting the terminal state $x_T=y$, the term ${g(t)}^2 h(x_t,t,y,T)$ pushes each forward step towards the end condition $y$, which exactly models the image degradation process~\citep[cf. Schr\"{o}dinger bridges,][]{de2021diffusion}. Correspondingly, the reverse-time SDE of Eq.~\eqref{eq:diffusion-bridge} can be written as
\begin{equation}
    \diff {x} = \Bigl[ f({x}, t) - g(t)^2\, \Bigl(s(x_t,t,y,T) - h(x_t,t,y,T) \Bigr) \Bigr] \diff t + g(t) \diff \hat{w},
    \label{eq:reverse-diffusion-bridge}
\end{equation}
where $s(x_t,t,y,T)=\nabla_{x_t} \log p(x_t \cond x_T) \cond_{x_T=y}$ is the conditional score function which can be learned via score-matching. The HQ image can then be recovered from the LQ image $y$ by iteratively running Eq.~\eqref{eq:reverse-diffusion-bridge} in time as a traditional SDE solver. Note that we can design specific SDEs (e.g., VP/VE-SDE~\cite{song2020score}) to make the function $h(x_t,t,y,T)$ tractable~\cite{zhou2023denoising,li2023bbdm,liu2023i2sb}. 
The simplest case is the Brownian bridge~\cite{li2023bbdm} which constructs the marginal distribution as $p(x_t|x_0, x_T)=\mathcal{N}((1-\frac{t}{T})x_0 + \frac{t}{T}x_T, \frac{t(T-t)}{T}I)$. Another particular case is the Schr{\"o}dinger bridge~\cite{liu2023i2sb}, which aims to compute a diffusion process that interpolates within the optimal coupling (when the reference measure of the bridge is chosen to be a Brownian motion) between the HQ and LQ image distributions~\cite{chen2021likelihood}. The solution of the Schr{\"o}dinger bridge converges weakly to an optimal transport plan w.r.t 2-Wasserstein~\citep{peyre2019computational,liu2023i2sb}.
Most diffusion bridge frameworks learn the noise $\epsilon_{\theta}(x_t, t)$ directly by adopting the similar score reparameterization trick from Eq.~\eqref{eq:score_to_noise}, which leads to the following objective: $\| \epsilon_{\theta}(x_t, t) - \frac{x_t - m_t}{\sqrt{v_t}} \|$, where $m_t$ and $v_t$ are the marginal mean and variance of the forward process.
More recently, \citet{yue2023image} further propose to apply the diffusion bridge to IR-SDE as the generalized Ornstein-Uhlenbeck bridge to achieve better performance. However, designing the forward SDE in Eq.~\eqref{eq:diffusion-bridge} with a tractable yet effective $h(x_t,t,y,T)$ remains a challenge and is under-explored in image restoration. With the growing popularity of Score-SDEs and diffusion bridges, we hope that future approaches will offer various efficient and elegant solutions to general image restoration problems.

\begin{figure}[t]
    \centering
    \includegraphics[width=1.\linewidth]{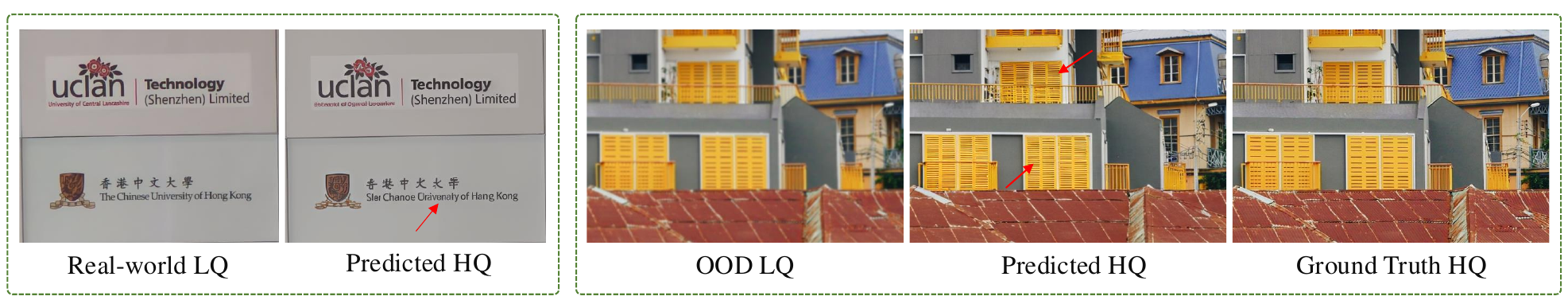}\vspace{-1.0mm}
    \caption{Failed examples of applying a trained diffusion model~\cite{yu2024scaling} on real-world and out-of-distribution (OOD) LQ input images. In the left example, the predicted HQ image contains unrecognizable text. In the right example, the generated window shutters are visually unpleasant and not consistent with the LQ input image.}\vspace{-1.0mm}
    \label{fig:ood}
\end{figure}

\section{Conclusion \& Discussion}\label{sec:discussion}

Diffusion models have shown incredible capabilities and gained significant popularity in generative modelling. In particular, the mathematics behind them make these models exceedingly elegant. Building on their core concepts, we described several approaches that effectively employ DMs for various image restoration tasks, achieving impressive results. However, it is also crucial to highlight the main challenges and further outline potential directions for future work. 

\begin{itemize}
    \item \textit{Difficult to process out-of-distribution (OOD) degradations:} Applying the trained DMs to OOD data often leads to inferior performance and produces visually unpleasant artifacts~\cite{luo2024photo}, as shown in Figure~\ref{fig:ood}. Some works~\cite{wang2024exploiting,lin2023diffbir} propose to address this issue by introducing the powerful Stable Diffusion (SD)~\cite{rombach2022high} with a feature control module~\cite{zhang2023adding}. Such approaches do however still need to refine the SD model with specific IR datasets. Moreover, the commonly used synthetic data strategy~\cite{wang2021real} just simulates known degradations such as noise, blur, compression, etc., and is unable to cover all corruption types which might be encountered in real-world applications. Inspired by the success of large language models and vision-language models, more recent approaches~\cite{luo2023controlling,luo2024photo,yu2024scaling,wu2024seesr} have begun to explore the use of various language-based image representations in IR. The main idea is to produce `clean' text descriptions of input LQ images, describing the main image content without undesired degradation-related concepts, and use these to guide the restoration process.

    \item \textit{Inconsistency in image generation:} While DMs produce photo-realistic results, the generated details are often inconsistent with the original input, especially regarding texture and text information, as shown in the right of Figure~\ref{fig:direct-diffusion} and in Figure~\ref{fig:ood}. This is mainly due to the intrinsic bias in the multi-step noise/score estimation and the stochasticity of the noise injection in each iteration. One solution is to add a predictor to generate the initial HQ image (with $\ell_1$ loss) and then gradually add more details via a diffusion process~\cite{whang2022deblurring}. However, this requires an additional network and the performance highly depends on the trained predictor. IR-SDE~\cite{luo2023image} proposes a maximum likelihood objective to learn the optimal restoration path, but its reverse-time process still contains noise injection (i.e. Wiener process) thus leading to unsatisfactory results. Recently, flow matching and optimal transport have shown great potential in image generation. In particular, they can form straight line trajectories in inference, which are more efficient than the curved paths of DMs~\cite{lipman2022flow,adrai2024deep}. Applying such methods to IR tasks is therefore a seemingly promising future direction.

    \item \textit{High computational cost and inference time:} Most diffusion-based image restoration methods require a significant number of diffusion steps to generate the final HQ image (typically 1\thinspace000 steps using DDPMs), which is both time-consuming and computationally costly, thus bringing challenges for deployment in various real-world applications. This problem can be alleviated using latent diffusion models (LDMs)~\cite{rombach2022high,luo2023refusion} or efficient sampling techniques~\cite{song2020denoising,lu2022dpm}. Unfortunately, these are not always suitable for IR tasks since the LDM often produces color shifting~\cite{wang2024exploiting}, and the efficient sampling would decrease the image generation quality~\cite{song2020denoising}. Considering the particularity of IR, several works~\citep{luo2023image,liu2023i2sb,corenflos2024conditioning} design the diffusion process towards degraded images (see~\ref{sec:diffusion_towards_lq}), such that their inference can start from the LQ image (rather than Gaussian noise). While this makes the sampling process more efficient (typically requiring less than 100 diffusion steps), it could be possible to improve further by designing more effective SDEs or diffusion bridge functions.
\end{itemize}

\paragraph{Closing} 
We have covered the basics of diffusion models and key techniques for applying them to IR tasks. This is an active research area with many interesting challenges and potential future directions, such as achieving photo-realistic yet consistent image generation, robustness to real-world image degradations, and more computationally efficient sampling. Ultimately, we hope this review paper offers a foundational understanding that enables readers to gain deeper insights into the mathematical principles underlying advanced diffusion-based IR approaches.

\vspace{0.5em} \noindent \textbf{Acknowledgements}
This research was partially supported by the \emph{Wallenberg AI, Autonomous Systems and Software Program (WASP)} funded by the Knut and Alice Wallenberg Foundation, by the project \emph{Deep Probabilistic Regression -- New Models and Learning Algorithms} (contract number: 2021-04301) funded by the Swedish Research Council, and by the \emph{Kjell \& M{\"a}rta Beijer Foundation}.


\bibliographystyle{plainnat}
\bibliography{main}

\end{document}